# Prediction with Advice of Unknown Number of Experts

**Alexey Chernov & Vladimir Vovk**
Computer Learning Research Centre
Department of Computer Science
Royal Holloway, University of London
Egham, Surrey TW20 0EX, UK
{chernov,vovk}@cs.rhul.ac.uk

## Abstract

In the framework of prediction with expert advice, we consider a recently introduced kind of regret bounds: the bounds that depend on the effective instead of nominal number of experts. In contrast to the Normal-Hedge bound, which mainly depends on the effective number of experts but also weakly depends on the nominal one, we obtain a bound that does not contain the nominal number of experts at all. We use the defensive forecasting method and introduce an application of defensive forecasting to multi-valued supermartingales.

## 1 INTRODUCTION

We consider the problem of prediction with expert advice (PEA) and its variant, decision-theoretic online learning (DTOL). In the PEA framework (see Cesa-Bianchi and Lugosi, 2006, for details, references and historical notes), at each step Learner gets decisions (also called predictions) of several Experts and must make his own decision. Then the environment generates an outcome and a (real-valued) loss is calculated for each decision as a known function of decision and outcome. The difference between the cumulative losses of Learner and one of Experts is the regret to this Expert. Learner aims at minimizing his regret to Experts, for any sequence of Expert decisions and outcomes.

In DTOL (Freund and Schapire, 1997) Learner's decision is a probability distribution on a finite set of actions. Then each action incurs a loss (the vector of the losses can be regarded as the outcome), and Learner suffers the loss equal to the expected loss over all actions (according to the probabilities from his decision). The regret is the difference between the cumulative losses of Learner and one of the actions. One can interpret each action as a rigid Expert that always suggests this action. A precise connection between the DTOL and PEA frameworks will be described in Section 2.

Usually Learner is required to have small regret to all Experts. In other words, a strategy for Learner must have a guaranteed upper bound on Learner's regret to the best Expert (one with the minimal loss). In this paper we deal with another kind of bound, recently introduced in Chaudhuri et al. (2009). It captures the following intuition. Generally speaking, the more Experts (or actions, in the DTOL terminology) Learner must take into account, the worse his performance will be. However, assume that each Expert has several different names, so Learner is given a lot of identical advice. It seems natural that the loss of Learner is big if there is a real controversy between Experts (or a real difference between actions), and small if most of the Experts agree with each other. So a competent regret bound should depend on the real number of Experts instead of the nominal one. Another example: assume that all the actions are different, but many of them are good — there are many ways to achieve some goal. Then Learner has less space to make a mistake and to select a bad action. Again it seems that a competent regret bound should depend on the fraction of the good actions rather than the nominal number of actions.

If the effective number of actions (Experts) is significantly less than the nominal one, one can loosely say that the number of actions is unknown in this setting. The following regret bound obtained by Chaudhuri et al. (2009) for their NormalHedge algorithm holds for this case:

$$L_T \leq L_T^\epsilon + O\left(\sqrt{T \ln \frac{1}{\epsilon}} + \ln^2 N\right), \qquad (1)$$

where $N$ is the nominal number of actions, $L_T$ is the cumulative loss of Learner after step $T$ and $L_T^\epsilon$ is any value such that at least $\epsilon$-fraction of actions have smaller or equal cumulative loss after step $T$ (or $L_T^\epsilon$ can be interpreted as the loss of $\epsilon N$th best action). It is important that the bound holds uniformly for all $\epsilon$ and $T$ and the algorithm does not need to know them in advance. The number $1/\epsilon$ plays the role of the *effective* number of actions. The bound shows, in a sense, that the NormalHedge algorithm can work even if the number of actions is not known.

Our main result (Theorem 3) is the following bound for a new algorithm:

$$L_T \leq L_T^\epsilon + 2\sqrt{T \ln \frac{1}{\epsilon}} + 7\sqrt{T}.$$

This bound is also uniform in $T$ and $\epsilon$. In contrast to (1), our bound does not depend on the nominal number of actions, whereas (1) contains a term $O(\ln^2 N)$. So it is the first (as far as we know) bound strictly in terms of the effective number of actions. Our bound has a simpler structure, but it is generally incomparable to the (precise) bound for Normal Hedge from Chaudhuri et al. (2009) (see Subsection 3.3 for discussion of different known bounds). Also our bound can be easily adapted to internal regret (see Khot and Ponnuswami, 2008, for definition). We omit this result as well as some proofs and explanations due to space constraints: see the full version of our paper (Chernov and Vovk, 2010).

Our bound is obtained with the help of the defensive forecasting method (DF). The DF is based on bounding the growth of some supermartingale (a kind of potential function). In Chernov et al. (2010), the DF was used to obtain bounds of the form $L_T \leq cL_T^n + a$, where $c$ and $a$ are some constants. For our form of bounds, we need a new variation of the DF and a new sort of supermartingales. So we introduce the notion of multivalued supermartingale and prove a boundedness result for them (Lemmas 5 and 6). (This result is of certain independent interest: for example, it helps to get rid of additional Assumption 3 in Theorem 20 in Chernov et al., 2010.)

The paper is organized as follows. In Section 2 we describe the setup of prediction with expert advice and of decision-theoretic online learning, and define the $\epsilon$-quantile regret. In Subsection 3.1 we introduce the Defensive Forecasting Algorithm, then in Subsection 3.2 we prove two bounds on the $\epsilon$-quantile regret, and in Subsection 3.3 we compare them with the bound for the NormalHedge algorithm and with other known bounds. In Section 4 we present technical details that we need to justify the algorithm: define multivalued supermartingales, discuss their properties, and introduce supermartingales of a specific form that are based on the Hoeffding inequality.

## 2 NOTATION AND SETUP

Vectors with coordinates $p_1, \ldots, p_N$ are denoted by an arrow over the letter: $\vec{p} = (p_1, \ldots, p_N)$. For any natural $N$, by $\Delta_N$ we denote the standard simplex in $\mathbb{R}^N$: $\Delta_N = \{\vec{p} \in [0,1]^N \mid \sum_{n=1}^N p_n = 1\}$. By $\vec{p} \cdot \vec{q}$ we denote the scalar product: $\vec{p} \cdot \vec{q} = \sum_{n=1}^N p_n q_n$.

---
**Protocol 1** Decision-Theoretic Online Learning
$L_0 := 0$.
$L_0^n := 0$, $n = 1, \ldots, N$.
**for** $t = 1, 2, \ldots$ **do**
  Learner announces $\vec{\gamma}_t \in \Delta_N$.
  Reality announces $\vec{\omega}_t \in [0,1]^N$.
  $L_t := L_{t-1} + \vec{\gamma}_t \cdot \vec{\omega}_t$.
  $L_t^n := L_{t-1}^n + \omega_{t,n}$, $n = 1, \ldots, N$.
**end for**

---

The decision-theoretic framework for online learning (DTOL) was introduced in Freund and Schapire (1997). The DTOL protocol is given as Protocol 1. The Learner has $N$ available actions, and at each step $t$ he must assign probability weights $\gamma_{t,1}, \ldots, \gamma_{t,N}$ to these actions. Then each action suffers a loss $\omega_{t,n}$, and Learner's loss is the expected loss over all actions according to the weights he assigned. Learner's goal is to keep small his regret $R_t^n = L_t - L_t^n$ to any action $n$, independent of the losses.

---
**Protocol 2** Prediction with Expert Advice
$L_0 := 0$.
$L_0^n := 0$, $n = 1, \ldots, N$.
**for** $t = 1, 2, \ldots$ **do**
  Expert $n$ announces $\gamma_t^n \in \Gamma$, $n = 1, \ldots, N$.
  Learner announces $\gamma_t \in \Gamma$.
  Reality announces $\omega_t \in \Omega$.
  $L_t := L_{t-1} + \lambda(\gamma_t, \omega_t)$.
  $L_t^n := L_{t-1}^n + \lambda(\gamma_t^n, \omega_t)$, $n = 1, \ldots, N$.
**end for**

---

DTOL can be regarded as a special case of prediction with expert advice (PEA), as explained below. The PEA protocol is given as Protocol 2. The game of prediction is specified by the set of outcomes $\Omega$, the set of decisions $\Gamma$ and the loss function $\lambda : \Gamma \times \Omega \to \mathbb{R}$. The game is played repeatedly by Learner having access to decisions made by a pool of Experts. At each step, Learner is given $N$ Experts' decisions and is required to come out with his own decision. The loss $\lambda(\gamma, \omega)$ measures the discrepancy between the decision $\gamma$ and the outcome $\omega$. $L_t$ is Learner's cumulative loss over the first $t$ steps, and $L_t^n$ is the $n$th Expert's cumulative loss over the first $t$ steps. The goal of Learner is the same: to keep small his regret $R_t^n = L_t - L_t^n$ to any Expert $n$, independent of Experts' moves and the outcomes.

As defined in Chaudhuri et al. (2009) (for DTOL), the *regret to the top $\epsilon$-quantile* (at step $T$) is the value $R_T^\epsilon$ such that there are at least $\epsilon N$ actions (the fraction at least $\epsilon$ of all Experts) with $R_T^n \geq R_T^\epsilon$. Or, equivalently, $R_T^\epsilon = L_T - L_T^\epsilon$ where $L_T^\epsilon$ is a value such that at least $\epsilon N$ actions (the fraction at least $\epsilon$ of all Experts) has the loss $L_T^n$ less than $L_T^\epsilon$.

A uniform bound on $R_T^\epsilon$ (in other words, a bound on Learner's loss $L_T$ in terms of $L_T^\epsilon$) that holds for all $\epsilon$ is more general than the standard best Expert bounds. The latter can be obtained as a special case for $\epsilon = 1/N$. For this reason, it is natural to call the value $1/\epsilon$ the *effective* number of actions: a bound on $R_T^\epsilon$ can be considered as the best Expert bound in an imaginary game against $1/\epsilon$ Experts.

Let us say what games $(\Omega, \Gamma, \Lambda)$ we consider in this paper. For any game $(\Omega, \Gamma, \lambda)$, we call $\Lambda = \{g \in \mathbb{R}^\Omega \mid \exists \gamma \in \Gamma \, \forall \omega \in \Omega \, g(\omega) = \lambda(\gamma, \omega)\}$ the *prediction set*. The prediction set captures most of the information about the game. The prediction set is assumed to be nonempty. In this paper, we consider *bounded convex compact games* only. This means that we assume that the set $\Lambda$ is bounded and compact, and the superprediction set $\Lambda + [0, \infty]^\Omega$ is convex, that is, for any $g_1, \ldots, g_K \in \Lambda$ and for any $p_1, \ldots, p_K \in [0,1]^K$, $\sum_{k=1}^K p_k = 1$, there exists $g \in \Lambda$ such that $g(\omega) \leq \sum_{k=1}^K p_k g_k(\omega)$ for all $\omega \in \Omega$. For such games, we assume without loss of generality that $\Lambda \subseteq [0,1]^\Omega$ (we can always scale the loss function).

For DTOL as a special case of PEA, the outcome space is $\Omega = [0,1]^N$, the decision space is $\Gamma = \Delta_N$, and the loss function is $\lambda(\vec{\gamma}, \vec{\omega}) = \vec{\gamma} \cdot \vec{\omega}$. Experts play fixed strategies always choosing $\vec{\gamma}_t^n$ such that $\gamma_{t,n}^n = 1$ and $\gamma_{t,k}^n = 0$ for $k \neq n$ (see e.g. Vovk, 1998, Example 7, for more details about this game).

In an important sense the general PEA protocol for the bounded convex games is equivalent to DTOL. Obviously, if some upper bound on regret is achievable in any PEA game then it is achievable in the special case of the DTOL game. To see how to transfer an upper bound from DTOL to a PEA game, let us interpret the decisions $\gamma_t^n$ of Experts and the outcome $\omega_t$ in the PEA game as the outcome $\vec{\omega}_t'$ in DTOL: $\omega_{t,n}' = \lambda(\gamma_t^n, \omega_t)$. If Learner's decision $\gamma_t$ satisfies $\lambda(\gamma_t, \omega_t) \leq \sum_{n=1}^N \gamma_{t,n}' \lambda(\gamma_t^n, \omega_t)$, where $\vec{\gamma}_t'$ is Learner's decision in DTOL, then the regret (at step $t$) in the PEA game will be not greater than the regret in DTOL. It remains to note that, since the game is convex, for any $\vec{\gamma}_t'$ there exists $\gamma_t$ such that $\lambda(\gamma_t, \omega) \leq \sum_{n=1}^N \gamma_{t,n}' \lambda(\gamma_t^n, \omega)$, for any $\omega \in \Omega$.

However, the equivalence between DTOL and PEA is limited. In particular, we can obtain PEA bounds that hold for specific loss functions or classes of loss functions, such as mixable loss functions (Vovk, 1998), and these bounds may be much stronger than the general bounds induced by DTOL.

In this paper, we consider PEA and DTOL in parallel for another reason. It is sometimes useful to consider a more general variant of Protocol 2 where the number of Experts is infinite (and maybe uncountably infinite): then PEA can be applied to large families of functions as Experts. With the help of our method, we can cope either with DTOL, where the number of actions is finite, or with PEA when $\Omega$ is finite and the number of Experts is arbitrary. So we cannot infer a bound for infinitely many Experts from a DTOL result, but we can obtain a PEA result directly. In the sequel, we will write about $N$ experts, but always allow $N$ to be infinite in the PEA case.

Most of the presentation below is in the terms of PEA but applicable to DTOL as well. We normally hide the difference between PEA and DTOL behind the common notation (DTOL is considered as the game described above). When the difference is important, we give two parallel fragments of a statement or a proof.

## 3 ALGORITHM AND BOUNDS

In this section, we formulate the algorithm and then obtain and discuss bounds for $\epsilon$-regret.

### 3.1 DEFENSIVE FORECASTING

The structure of the *Defensive Forecasting Algorithm* (DFA) is quite simple and is shown as Algorithm 1.

---
**Algorithm 1** Defensive Forecasting Algorithm
---
1: **for** $t = 1, 2, \ldots$ **do**
2:    Get Experts' predictions $\gamma_t^n \in \Gamma$, $n = 1, \ldots, N$.
3:    Find $\gamma \in \Gamma$ s.t. for all $\omega \in \Omega$
4:        $f_t(\gamma, \omega) \leq f_{t-1}(\gamma_{t-1}, \omega_{t-1})$.
5:    Output $\gamma_t := \gamma$.
6:    Get $\omega_t \in \Omega$.
7: **end for**.
---

The only meaningful step of Algorithm 1 is in lines 3–4. A function $f_t : \Gamma \times \Omega \to \mathbb{R}$ in line 4 can depend on all the previous Experts and Learner decisions, on the previous outcomes, and on Experts' decisions at this step (at the first step $f_0$ in the right-hand side is some constant). The sequence of functions $f_t$ is chosen in such a way that the inequality in lines 3–4 always has a solution and the inequality $f_t(\gamma_t, \omega_t) \leq f_0$ implies the loss bound we would like to obtain. The latter inequality follows from the fact that at each step $t$ the algorithm guarantees $f_t(\gamma_t, \omega_t) \leq f_{t-1}(\gamma_{t-1}, \omega_{t-1})$.

To prove that the inequality in lines 3–4 for our choice of $f_t$ is always soluble we use the notion of supermartingale, which is explained in Section 4. Some supermartingales were introduced and analyzed in Chernov et al. (2010). Below we use a new variety, which we call Hoeffding supermartingales. However, the theory of supermartingales is not so relevant for implementation of the algorithm. To solve the inequality in lines 3–4, an appropriate numerical method should be used. We do not consider practical issues in this paper and concentrate on theory.

Let us begin with a simple theorem that shows a clean application of the DFA.

**Theorem 1.** *If $T$ is known in advance then the DFA achieves the bound*
$$L_T \le \min_n L_T^n + \sqrt{2T \ln N}$$
*(for DTOL with $N$ actions as well as for PEA with $N$ experts).*

*Proof.* Let $\eta = \sqrt{2(\ln N)/T}$ and
$$f_t(\gamma, \omega) = \sum_{n=1}^N \frac{1}{N} e^{\eta(L_{t-1} - L_{t-1}^n) - \eta^2/2} \\ \times e^{\eta(\lambda(\gamma,\omega) - \lambda(\gamma_t^n, \omega)) - \eta^2/2}. \quad (2)$$

At each step $t$, Algorithm 1 in lines 3–4 finds $\gamma_t$ such that $f_t(\gamma_t, \omega) \le f_{t-1}(\gamma_{t-1}, \omega_{t-1})$ for all $\omega \in \Omega$. The existence of such $\gamma_t$ follows from the supermartingale results below, namely, Lemma 9 combined with Lemma 6 for DTOL or Lemma 5 for PEA. Clearly, $f_T(\gamma_t, \omega_t) = \sum_{n=1}^N \frac{1}{N} \exp(\eta(L_t - L_t^n) - \eta^2/2)$, and we get that the DFA applied to the sequence $\{f_t\}_{t=1}^T$ guarantees that
$$f_t(\gamma_T, \omega_T) = \sum_{n=1}^N \frac{1}{N} e^{\eta(L_T - L_T^n) - \eta^2/2} \le 1.$$
Bounding the sum from below by one addend, we get the bound. □

This bound is twice as large as the optimal bound obtained by Cesa-Bianchi and Lugosi (2006); see their Theorems 2.2 and 3.7. Our bound is the same as that in their Corollary 2.2, and the algorithm is not really different (the exponentially weighted average forecaster used there can be considered as the DFA with a certain solution of the inequality in lines 3–4).

### 3.2 BOUNDS ON $\epsilon$-QUANTILE REGRET

The bound in Theorem 1 is guaranteed only once, at step $T$ specified in advance. The next bound is uniform, that is, holds for any $T$, and also it holds for $\epsilon$-quantile regret for all $\epsilon > 0$.

**Theorem 2.** *For DTOL with $N$ actions, the DFA achieves the bound*
$$\int_0^{1/e} e^{(L_T - L_T^\epsilon)\eta - T\eta^2/2} \frac{d\eta}{\eta \left(\ln \frac{1}{\eta}\right)^2} \le \frac{1}{\epsilon}, \quad (3)$$
*for any $T$ and any $\epsilon > 0$, where $L_T^\epsilon$ is any value such that at least $\epsilon$-fraction of actions has the loss after step $T$ not greater than $L_T^\epsilon$. In particular, (3) implies*
$$L_T \le L_T^\epsilon + \left(1 + \frac{1}{\ln T}\right) \sqrt{2T \ln \frac{1}{\epsilon} + 5T \ln \ln T} \\ + O\left(\ln \frac{1}{\epsilon}\right). \quad (4)$$

*The bound holds also for PEA; if each of finitely or infinitely many Experts is assigned some positive weight $p_n$, the sum of all weights being not greater than 1, the DFA achieves (3) and (4) with $L_T^\epsilon$ being any value such that the total weight of Experts that have the loss after step $T$ not greater than $L_T^\epsilon$ is at least $\epsilon$.*

*Proof.* The idea is to mix all the supermartingales (the functions $f_t$) used in (2) over $\eta \in [0, 1/e]$ according to the probability measure $\mu(d\eta) = d\eta \Big/ \left(\eta (\ln(1/\eta))^2\right)$, $\eta \in [0, 1/e]$. That is, we apply Algorithm 1 taking in line 4 the following functions $f_t$:
$$f_t(\gamma, \omega) = \sum_{n=1}^N \frac{1}{N} \int_0^{1/e} \frac{d\eta}{\eta \left(\ln \frac{1}{\eta}\right)^2} e^{\eta(L_{t-1} - L_{t-1}^n) - \eta^2/2} \\ \times e^{\eta(\lambda(\gamma,\omega) - \lambda(\gamma_t^n, \omega)) - \eta^2/2} \quad (5)$$

(for PEA with weighted Experts, the term $1/N$ is replaced by $p_n$). Again, Lemma 9 combined with Lemma 6 for DTOL or Lemma 5 for PEA guarantees that the inequality in lines 3–4 is soluble. Thus, the algorithm achieves $f_T(\gamma_T, \omega_T) \le 1$ for all $T$. Bounding the sum from below by the sum of terms where $L_T^n \le L_T^\epsilon$, we get
$$\int_0^{1/e} e^{\eta(L_T - L_T^\epsilon) - T\eta^2/2} \frac{d\eta}{\eta \left(\ln \frac{1}{\eta}\right)^2} \le \frac{1}{\epsilon}.$$

Lower-bounding the integral, we get bound (4). □

Bounds uniform in $T$ are usually obtained with the help of $\eta = \eta_t$ changing at each step. In Theorem 2 we use a different method: the integral over $\eta$ of supermartingales informally correspond to running algorithms for all $\eta$ in parallel and taking some weighted average of their predictions.

The bound (4) has suboptimal asymptotics in $T$: the regret term growth rate is $O(\sqrt{T \ln \ln T})$ as $T \to \infty$,

instead of $O(\sqrt{T})$. The next theorem gives a bound with the optimal growth rate but using a "fake" DFA and time-varying $\eta_t \propto \sqrt{t}$.

**Theorem 3.** *For DTOL with $N$ actions, there exists a strategy that achieves the bound*

$$L_T \le L_T^\epsilon + 2\sqrt{T \ln \frac{1}{\epsilon}} + 7\sqrt{T} \qquad (6)$$

*for any $T$ and any $\epsilon > 0$, where $L_T^\epsilon$ is any value such that at least $\epsilon$-fraction of actions has the loss after step $T$ not greater than $L_T^\epsilon$.*

*The bound holds also for PEA; if each of finitely or infinitely many Experts is assigned some positive weight $p_n$, the sum of all weights being not greater than 1, the strategy achieves (6) with $L_T^\epsilon$ being any value such that the total weight of Experts that have the loss after step $T$ not greater than $L_T^\epsilon$ is at least $\epsilon$.*

*Proof.* The algorithm in this theorem differs from Algorithm 1: line 4 is replaced by $f_t(\gamma_t, \omega_t) \le C_t$, where $f_t$ is defined by (7) and $C_t$ by (8). This is not a proper use of supermartingales: the values $f_t(\gamma_t, \omega_t)$ may increase at some steps and $f_t(\gamma_t, \omega_t) \le f_{t-1}(\gamma_{t-1}, \omega_{t-1})$ does not hold. Nevertheless, the increases of $f_t$ stay bounded so that $f_t(\gamma_t, \omega_t) \le 1$ always holds.

Let $1/c = \sum_{i=1}^\infty (1/i^2)$. At step $T$, our algorithm finds $\gamma_T$ such that $f_T(\gamma_T, \omega) \le C_T$ for all $\omega$, where

$$\begin{aligned} f_T(\gamma, \omega) = & \sum_{n=1}^N \frac{1}{N} \sum_{i=1}^\infty \frac{c}{i^2} \\ & \times e^{(i/\sqrt{T})(L_{T-1} - L_{T-1}^n) - (i/2\sqrt{T}) \sum_{t=1}^{T-1}(i/\sqrt{t})} \\ & \times e^{(i/\sqrt{T})(\lambda(\gamma,\omega) - \lambda(\gamma_T^n,\omega)) - (i/\sqrt{T})^2/2} \end{aligned} \qquad (7)$$

and

$$\begin{aligned} C_T = & \sum_{n=1}^N \frac{1}{N} \sum_{i=1}^\infty \frac{c}{i^2} \\ & \times e^{(i/\sqrt{T})(L_{T-1} - L_{T-1}^n) - (i/2\sqrt{T}) \sum_{t=1}^{T-1}(i/\sqrt{t})}. \end{aligned} \qquad (8)$$

For PEA with weighted experts, it is sufficient to replace $1/N$ by $p_n$ in the definitions of $f_T$ and $C_T$.

Note that $f_T$ has the form (17): $f_T = \sum_{k=1}^K p_{T,k} H_{T,k}$, and $C_T = \sum_{k=1}^K p_{T,k}$. Hence Lemma 9 applies, and due to Lemma 6 or Lemma 5 such a $\gamma_T$ exists.

Let us prove by induction over $T$ that $C_T \le 1$. It is trivial for $T = 0$, since $L_0 = L_0^n = 0$ and $\sum_{t=1}^0 = 0$. Assume that $C_T \le 1$ and prove that $C_{T+1} \le 1$. By the choice of $\gamma_T$, we know that $f_T(\gamma_T, \omega_T) \le C_T \le 1$. Since the function $x^\alpha$ is concave for $0 < \alpha < 1$, we have $1 \ge (f_T(\gamma_T, \omega_T))^{\sqrt{T}/\sqrt{T+1}} \ge C_{T+1}$.

Now it is easy to get the loss bound. Assume that for an $\epsilon$-fraction of Experts their losses $L_T^n$ are smaller than or equal to $L_T^\epsilon$. Then $f_T(\gamma_T, \omega_T)$ can be bounded from below by

$$\epsilon \sum_{i=1}^\infty \frac{c}{i^2} e^{(i/\sqrt{T})(L_T - L_T^\epsilon) - (i/2\sqrt{T}) \sum_{t=1}^T (i/\sqrt{t})}.$$

Bounding the infinite sum by the term with $i = \lceil \sqrt{\ln(1/\epsilon)} \rceil + 1$, and using $\sum_{t=1}^T (1/\sqrt{t}) \le 2\sqrt{T}$, we obtain the final bound. □

**Remark 1.** For DTOL and for PEA with a finite number of Experts, the infinite sum over $i$ in the proof can be replaced by the sum up to $\lceil \sqrt{\ln N} \rceil + 1$. However, one should keep decreasing weights $c/i^2$: for uniform weights the bound will have an additional term of the form $O((\ln \ln N)/\ln(1/\epsilon))$.

**Remark 2.** Probably, the first bound for $\epsilon$-quantile regret was stated (implicitly) in Freund et al. (1997). More precisely, that paper considered even more general regret notion: Theorem 1 in Freund et al. (1997) gives a bound for PEA with weighted experts under the logarithmic loss of the form

$$L_T \le \sum_{n=1}^N u_n L_T^n + \sum_{n=1}^N u_n \ln \frac{u_n}{p_n}$$

for any $\vec{u} \in \Delta_N$; $p_1, \ldots, p_N$ are weights of Experts. Here $p_n$ are known to the algorithm in advance, whereas $u_n$ are not known and the bound holds uniformly for all $u_n$. Taking $u_n = 0$ for Experts not from the $\epsilon$-quantile of the best Experts, and uniform $u_n$ over Experts from the $\epsilon$-quantile, we get the bound in terms of $L_T^\epsilon$. It can be easily checked that the strategy in Theorem 3 also achieves the following bound:

$$L_T \le \sum_{n=1}^N u_n L_T^n + 2\sqrt{T \left(\sum_{n=1}^N u_n \ln \frac{u_n}{p_n}\right)} + 7\sqrt{T}$$

for any $\vec{u} \in \Delta_N$ and any $T$. In Theorem 2 one can replace $L_T^\epsilon$ by $\sum_{n=1}^N u_n L_T^n$ and $\ln(1/\epsilon)$ by $\sum_{n=1}^N u_n \ln(u_n/p_n)$ as well.

### 3.3 DISCUSSION OF BOUNDS

For a game with $N$ Experts, the best bound that is uniform in $T$ is given by Cesa-Bianchi and Lugosi (2006), Theorem 2.3:

$$L_T \le L_T^n + \sqrt{2T \ln N} + \sqrt{\frac{\ln N}{8}}. \qquad (9)$$

The bounds (4) and (6) with $\epsilon = 1/N$ are always worse than (9). However, it appears that the bound (9) cannot be transferred to $\epsilon$-quantile regret $R_T^\epsilon = L_T - L_T^\epsilon$.

The proof of Theorem 2.3 in Cesa-Bianchi and Lugosi (2006) heavily relies on tracking the loss of only one best Expert, and it is unclear whether the existence of several good (or identical) Experts can be exploited in this proof. The experiments reported in Chaudhuri et al. (2009) show that algorithms with good best Expert bounds may have rather bad performance when the nominal number of Experts is much greater than the effective number of Experts.

The first (and the only, as far as we know) bound specifically formulated for $\epsilon$-quantile regret is proven for the NormalHedge algorithm by Chaudhuri et al. (2009), Theorem 1:

$$L_T \leq L_T^\epsilon + \sqrt{1 + \ln \frac{1}{\epsilon}}$$
$$\times \sqrt{3(1+50\delta)T + \frac{16\ln^2 N}{\delta}\left(\frac{10.2}{\delta^2} + \ln N\right)}, \quad (10)$$

which holds uniformly for all $\delta \in (0, 1/2]$. Note that this bound depends on the effective number of actions $1/\epsilon$ and at the same time on the nominal number of actions $N$. The latter dependence is weak, but probably prevents the use of NormalHedge with infinitely many Experts.

The main advantage of our bounds in Theorems 2 and 3 is that they are stated exclusively in terms of the effective number of Experts. In a sense, the DFA does not need to know in advance the number of Experts. (To obtain a precise statement of this kind, one can consider the setting where Experts may come at some later steps; the regret to a late Expert is accumulated over the steps after his coming. Our algorithms and bounds can be easily adapted for this setting.)

Both our bounds are asymptotically worse than (10) when $\epsilon$ and $N$ are fixed and $T \to \infty$. In this case, the regret term in (10) grows as $\sqrt{3T\ln(1/\epsilon)} + 3T$, whereas in (6) it grows as $\sqrt{4T\ln(1/\epsilon)} + 7\sqrt{T}$ (faster by a constant factor) and in (4) it grows as $\sqrt{5T\ln\ln T + 2T\ln(1/\epsilon)}$ (faster by $\ln\ln T$).

On the other hand, our bounds are better when $T$ is relatively small. The term $\ln\ln T$ is small for any reasonable practical application (e.g., $\ln\ln T < 4$ if $T$ is the age of the universe expressed in microseconds), and then the main term in (4) is $\sqrt{2T\ln(1/\epsilon)}$, which even fits the optimal bound (9). Bound (6) improves over (10) for $T \leq 10^6 \ln^4 N$.

Now let us say a few words about known algorithms for which $\epsilon$-quantile regret bounds were not formulated explicitly but can easily be obtained.

The Weighted Average Algorithm, which is used to obtain bound (9), can be analysed in a manner different from the proof of Theorem 2.3 in Cesa-Bianchi and Lugosi (2006): see Kalnishkan and Vyugin (2005). Then one can obtain the following bound for the $\epsilon$-quantile regret:

$$L_T \leq L_T^\epsilon + \frac{1}{c}\sqrt{T}\ln\frac{1}{\epsilon} + c\sqrt{T},$$

where the constant $c > 0$ is arbitrary but must be fixed in advance. If $\epsilon$ is not known and hence $c$ cannot be adapted to $\epsilon$, the leading term is $O(\sqrt{T}\ln\frac{1}{\epsilon})$, which is worse than (6) for small $\epsilon$ (that is, if we consider a large effective number of actions).

For the Aggregating Algorithm (Vovk, 1998) (which can be considered as a special case of the DFA for a certain supermartingale, as shown in Chernov et al., 2010), the bound can be trivially adapted to $\epsilon$-quantile regret:

$$L_T \leq cL_T^\epsilon + \frac{c}{\eta}\ln\frac{1}{\epsilon},$$

where the possible constants $c \geq 1$ and $\eta > 0$ depend on the loss function. However, in the case of DTOL or arbitrary convex games, the constant $c$ is strictly greater that 1 and the bound may be much worse than (4) and (6) (when $L_T^\epsilon$ grows significantly faster than $\sqrt{T}$). At the same time, this bound is much better when $L_T^\epsilon \approx 0$ (there is at least $\epsilon$ fraction of "perfect" Experts).

For the standard setting with the known number of Experts, other "small loss" bounds, of the form $L_T \leq L_T^n + O(\sqrt{L_T^n})$, have been obtained. Chaudhuri et al. (2009) posed an open question whether similar bounds can be obtained if the (effective) number of actions is not known. We left the question open.

## 4 SUPERMARTINGALES

In this section we present the technical results that allow us to prove that Algorithm 1 (and the "fake" version of the DFA from Theorem 3) can always find $\gamma$ in line 4.

Let $\Omega$ be a compact metric space. Any finite set $\Omega$ is considered as a metric space with the discrete metric. Let $\mathcal{P}(\Omega)$ be the space of all measures on $\Omega$ supplied with the weak topology.

For any measurable function $g \in \mathbb{R}^\Omega$ and for any $\pi \in \mathcal{P}(\Omega)$, denote

$$\mathrm{E}_\pi g = \int_\Omega g(\omega)\pi(d\omega).$$

For finite $\Omega$, this reduces to the scalar product:

$$\mathrm{E}_\pi g = \sum_{\omega \in \Omega} g(\omega)\pi(\{\omega\}).$$

Let $S$ be an operator that to any sequence $e_1, \pi_1, \omega_1, \ldots, e_{T-1}, \pi_{T-1}, \omega_{T-1}, e_T$, where $\omega_t \in \Omega$, $\pi_t \in \mathcal{P}(\Omega)$, $t = 1, \ldots, T-1$, and $e_t$, $t = 1, \ldots, T$ are some arbitrary values, assigns a function $S_T : \mathcal{P}(\Omega) \to \mathbb{R}^\Omega$. To simplify notation, we will hide the dependence of $S_T$ on all the long argument sequence in the index $T$. We call $S$ a (game-theoretic) supermartingale if for any sequence of arguments, for any $\pi \in \mathcal{P}(\Omega)$, for $g_{T-1} = S_{T-1}(\pi_{T-1})$ and for $g = S_T(\pi)$ it holds

$$\mathrm{E}_\pi g \leq g_{T-1}(\omega_{T-1}). \tag{11}$$

This definition of supermartingale is equivalent to the one given in Chernov et al. (2010). We say that supermartingale $S$ is forecast-continuous if every $S_T$ is a continuous function.

The main property of forecast-continuous supermartingales that makes them useful in our context is given by Lemma 4. Originally, a variant of the lemma was obtained by Leonid Levin in 1976. The proof is based on fixed-point considerations; see Theorem 6 in Gács (2005) or Lemma 8 in Chernov et al. (2010) for details.

**Lemma 4.** *Let $\Omega$ be a compact metric space. Let a function $q : \mathcal{P}(\Omega) \times \Omega \to \mathbb{R}$ be continuous as function from $\mathcal{P}(\Omega)$ to $\mathbb{R}^\Omega$. If for all $\pi \in \mathcal{P}(\Omega)$ it holds that*

$$\mathrm{E}_\pi q(\pi, \cdot) \leq C,$$

*where $C \in \mathbb{R}$ is some constant, then*

$$\exists \pi \in \mathcal{P}(\Omega) \, \forall \omega \in \Omega \quad q(\pi, \omega) \leq C.$$

The lemma guarantees that for any forecast-continuous supermartingale $S$ we can always choose $g_t \in S_t$ such that $g_t(\omega) \leq g_{t-1}(\omega_{t-1})$ for all $\omega$. This is exactly the kind of condition we need for the DFA.

Unfortunately, for the loss bounds we want to obtain, we did not find a suitable forecast-continuous supermartingale. So we define a more general notion of multivalued supermartingale, and prove an appropriate variant of Levin's lemma.

### 4.1 MULTIVALUED SUPERMARTINGALES

To get the definition of a multivalued supermartingale, we make just three changes in the definition of supermartingale above: $S_T$ is function from $\mathcal{P}(\Omega)$ to non-empty subsets of $\mathbb{R}^\Omega$; operator $S$ depends additionally on $g_t \in S_t(\pi_t)$; the condition (11) holds for any $g \in S_T(\pi)$. Namely, let $S$ be an operator that to any sequence $e_1, \pi_1, g_1, \omega_1, \ldots, e_{T-1}, \pi_{T-1}, g_{T-1}, \omega_{T-1}, e_T$, where $\omega_t \in \Omega$, $\pi_t \in \mathcal{P}(\Omega)$, $g_t \in \mathbb{R}^\Omega$, $t = 1, \ldots, T-1$, and $e_t$, $t = 1, \ldots, T$, are some arbitrary values, assigns a function $S_T : \mathcal{P}(\Omega) \to 2^{\mathbb{R}^\Omega}$ such that $S_T(\pi)$ is a non-empty subset of $\mathbb{R}^\Omega$ for all $\pi \in \mathcal{P}(\Omega)$. $S$ is called a *multivalued supermartingale* if for any sequence of arguments where $g_t \in S_t(\pi_t)$, for any $\pi \in \mathcal{P}(\Omega)$, $S_T(\pi) \neq \emptyset$ and for all $g \in S_T(\pi)$ it holds

$$\mathrm{E}_\pi g \leq g_{T-1}(\omega_{T-1}). \tag{12}$$

A multivalued supermartingale is called *forecast-continuous* if for every $S_T$, the set $\{(\pi, g) \mid \pi \in \mathcal{P}(\Omega), g \in S_T(\pi)\}$ is closed and additionally for every $\pi \in \mathcal{P}(\Omega)$ the set $S_T(\pi) + [0, \infty]^\Omega = \{g \in \mathbb{R}^\Omega \mid \exists g' \in S_T(\pi) \forall \omega \in \Omega \; g'(\omega) \leq g(\omega)\}$ is convex.

Note that if $S$ is a forecast-continuous multivalued supermartingale and $S_t(\pi)$ always consists of exactly one element, $S$ is (equivalent to) a forecast-continuous supermartingale in the former sense: the graph of $S_T$ being closed means that $S_T(\pi)$ is a continuous function of $\pi$ and the convexity requirement becomes trivial.

### 4.2 LEVIN LEMMA FOR MULTIVALUED SUPERMARTINGALES

Here we prove two versions of Levin's lemma suitable for multivalued supermartingales. The first variant (it is simpler) is used for PEA with finite outcome set $\Omega$. The second variant is used for DTOL.

**Lemma 5.** *Let $\Omega$ be a finite set. Let $X$ be a compact subset of $\mathbb{R}^\Omega$. Let $q \subseteq \mathcal{P}(\Omega) \times X$ be a relation. Denote $q(\pi) = \{g \mid (\pi, g) \in q\}$ and $\mathrm{ran}\, q = \cup_{\pi \in \mathcal{P}(\Omega)} q(\pi) \subseteq X$. Suppose that $q$ is closed, for every $\pi \in \mathcal{P}(\Omega)$ the set $q(\pi)$ is non-empty and the set $q(\pi) + [0, \infty]^\Omega$ is convex. If for some real constant $C$ it holds that for every $\pi \in \mathcal{P}(\Omega)$*

$$\forall g \in q(\pi) \quad \mathrm{E}_\pi g \leq C,$$

*then there exists $g \in \mathrm{ran}\, q$ such that*

$$\forall \omega \in \Omega \quad g(\omega) \leq C.$$

We derive the lemma from Lemma 4 similarly to the derivation of Kakutani's fixed point theorem for multivalued mappings (see, e. g. Agarwal et al., 2001, Theorem 11.9) from Brouwer's fixed point theorem. Unfortunately, we did not find a way just to refer to Kakutani's theorem and have to repeat the whole construction with appropriate changes.

*Proof.* Note first that $\mathcal{P}(\Omega)$ is compact for finite $\Omega$, hence $q$ is compact as a closed subset of a compact set. Let $M_q = \max_{g \in \mathrm{ran}\, q, \omega \in \Omega} |g(\omega)|$.

For every natural $m > 0$, let us take any $(1/m)$-net $\{\pi_k^m\}$ on $\mathcal{P}(\Omega)$ such that for every $\pi \in \mathcal{P}(\Omega)$ there is at least one net element $\pi_k^m$ at the distance less than $1/m$ from $\pi$ and for every $\pi \in \mathcal{P}(\Omega)$ there are at most

$4|\Omega|^2$ elements of the net at the distances less than $1/m$ from $\pi$. (One can use here any reasonable distance on $\mathcal{P}(\Omega)$, for example, the maximum absolute value of the coordinates of the difference.) For every $\pi_k^m$ in the net, fix any $g_k^m \in q(\pi_k^m)$ (recall that $q(\pi_k^m)$ is non-empty).

Now let us define a function $q^m : \mathcal{P}(\Omega) \times \Omega \to \mathbb{R}$ as a linear interpolation of the points $(\pi_k^m, g_k^m)$. Namely, let $\{u_k^m\}$ be a partition of unity of $\mathcal{P}(\Omega)$ subordinate to $U_{1/m}(\pi_k^m)$, the $(1/m)$-neighborhoods of $\pi_k^m$ (that is, $u_k^m(\pi)$ are non-negative, $u_k^m(\pi) = 0$ if the distance between $\pi$ and $\pi_k^m$ is $1/m$ or more, and the sum over $k$ of all $u_k^m(\pi)$ is 1 at any $\pi$). Let $q^m(\pi, \omega) = \sum_k u_k^m(\pi) g_k^m(\omega)$.

Clearly, the function $q^m$ is forecast-continuous. Let us find an upper bound on its expectation: $\mathrm{E}_\pi q^m(\pi, \cdot) = \sum_k u_k^m(\pi) \mathrm{E}_\pi g_k^m = \sum_k u_k^m(\pi) \mathrm{E}_{\pi_k^m} g_k^m + \sum_k u_k^m(\pi) \sum_{\omega \in \Omega} (\pi(\{\omega\}) - \pi_k^m(\{\omega\})) g_k^m(\omega) \leq C + M_q |\Omega|/m$ (the bound on the first term holds since $g_k^m \in q(\pi_k^m)$ and hence $\mathrm{E}_{\pi_k^m} g_k^m \leq C$).

By Lemma 4 we can find a point $\pi^m \in \mathcal{P}(\Omega)$ such that

$$\forall \omega \in \Omega \quad q^m(\pi^m, \omega) \leq C + M_q |\Omega|/m \,.$$

Recalling that $q^m(\pi^m, \omega) = \sum_k u_k^m(\pi^m) g_k^m(\omega)$ and that there are at most $4|\Omega|^2$ non-zero values among $u_k^m(\pi^m)$, we get the following statement. There exist some $\alpha_k^m \geq 0$, $k = 1, \ldots, 4|\Omega|^2$, $\sum_k \alpha_k^m = 1$, and some $g_k^m \in q(\pi_k^m)$ with $\pi_k^m$ at the distance at most $1/m$ from $\pi^m$ such that

$$\forall \omega \in \Omega \quad \sum_{k=1}^{4|\Omega|^2} \alpha_k^m g_k^m(\omega) \leq C + M_q |\Omega|/m \,. \quad (13)$$

Since $\mathcal{P}(\Omega)$ is compact, we can find a limit point $\pi^*$ of $\pi^m$. It will be a limit point of $\pi_k^m$ as well. Since $q$ is compact, we can find $g_k^* \in q(\pi^*)$ such that $(\pi^*, g_k^*)$ are limit points of $(\pi_k^m, g_k^m)$ for each $k$. Finally, since $\mathcal{P}(\{1, \ldots, 4|\Omega|^2\})$ is compact, we can find limit points $\alpha_k^*$ (corresponding to the points $g_k^*$).

Taking the limits as $m \to \infty$ over the convergent subsequences in (13), we get

$$\forall \omega \in \Omega \quad \sum_{k=1}^{4|\Omega|^2} \alpha_k^* g_k^*(\omega) \leq C \,.$$

Since $q(\pi^*) + [0, \infty]^\Omega$ is convex, the convex combination $\sum_{k=1}^{4|\Omega|^2} \alpha_k^* g_k^*$ belongs to $q(\pi^*) + [0, \infty]^\Omega$. In other words, the combination is minorized by some $g^* \in q(\pi^*)$ and

$$g^*(\omega) \leq \sum_{k=1}^{4|\Omega|^2} \alpha_k^* g_k^*(\omega) \leq C$$

for all $\omega \in \Omega$. □

Now let us state a variant of the lemma suitable for the DTOL framework, where the set of outcomes is infinite. Here we make a strong assumption: the supermartingale values $S_T(\pi)$ depend on $\pi$ in a very limited way: just on the mean of $\pi$.

**Lemma 6.** *Let $\Omega$ be $[0, 1]^N$. Let $X$ be a compact subset of $\mathbb{R}^\Omega$. Let $q \subseteq \mathcal{P}(\Omega) \times X$ be a relation. Denote $q(\pi) = \{g \mid (\pi, g) \in q\}$ and $\operatorname{ran} q = \cup_{\pi \in \mathcal{P}(\Omega)} q(\pi) \subseteq X$. Assume that if $\int \omega \pi_1(d\omega) = \int \omega \pi_2(d\omega)$ then $q(\pi_1) = q(\pi_2)$. Suppose that $q$ is closed, for every $\pi \in \mathcal{P}(\Omega)$ the set $q(\pi)$ is non-empty and the set $q(\pi) + [0, \infty]^\Omega$ is convex. If for some real constant $C$ it holds that for every $\pi \in \mathcal{P}(\Omega)$*

$$\forall g \in q(\pi) \quad \mathrm{E}_\pi g \leq C \,,$$

*then there exists $g \in \operatorname{ran} q$ such that*

$$\forall \omega \in \Omega \quad g(\omega) \leq C \,.$$

The proof is very similar to the proof of Lemma 5. We need an additional step, since the compact space $\mathcal{P}([0, 1]^N)$ is infinite-dimensional, whereas we need a fixed number of terms in linear combinations. To guarantee the latter, we consider net elements $\pi_k^m$ with certain expected values only, and group the functions $g_k^m$ that correspond to the same expected value (which is possible due to the additional requirement in the lemma statement). See Chernov and Vovk (2010) for details.

### 4.3 HOEFFDING SUPERMARTINGALES

Here we introduce a specific multivalued supermartingale, or rather a family of supermartingales, that is used in our main results.

For technical convenience, our definition of supermartingale $S_t$ consists of two parts: a function $G : \mathcal{P}(\Omega) \to 2^\Gamma$, which assigns a set of decisions $G(\pi) \subseteq \Gamma$ to every $\pi \in \mathcal{P}(\Omega)$, and a function $f_t : \Gamma \times \Omega \to \mathbb{R}$. The values of $S_t$ are defined by the formula:

$$S_t(\pi) = \{g \in \mathbb{R}^\Omega \mid \exists \gamma \in G(\pi) \forall \omega \; g(\omega) = f_t(\gamma, \omega)\} \,. \quad (14)$$

The part $G(\pi)$ depends on the game $(\Omega, \Gamma, \lambda)$ only and does not change from step to step:

$$G(\pi) = \arg\min_{\gamma \in \Gamma} \mathrm{E}_\pi \lambda(\gamma, \cdot) =$$
$$\{\gamma \in \Gamma \mid \forall \gamma' \in \Gamma \; \mathrm{E}_\pi \lambda(\gamma, \cdot) \leq \mathrm{E}_\pi \lambda(\gamma', \cdot)\} \,. \quad (15)$$

**Lemma 7.** *Let $(\Omega, \Gamma, \lambda)$ be a game such that its prediction set $\Lambda = \{g \in \mathbb{R}^\Omega \mid \exists \gamma \in \Gamma \forall \omega \in \Omega \; g(\omega) = \lambda(\gamma, \omega)\}$ is a non-empty compact subset of $\mathbb{R}^\Omega$ and $\Lambda + [0, \infty]^\Omega$ is convex. Then the set*

$$G_\Lambda = \{(\pi, g) \in \mathcal{P}(\Omega) \times \Lambda \mid \exists \gamma \in G(\pi) \forall \omega \; g(\omega) = \lambda(\gamma, \omega)\}$$

is closed and for every $\pi \in \mathcal{P}(\Omega)$ the sets $G(\pi)$ and $G_\Lambda(\pi) = \{g \mid (\pi, g) \in G_\Lambda\}$ are non-empty and the sets $G_\Lambda(\pi) + [0, \infty]^\Omega$ are convex.

We omit all proofs in this subsection due to space constraints; see Chernov and Vovk (2010).

Note that for convex bounded compact games the conditions of the lemma are satisfied by definition. For DTOL, the set $\Lambda = \{g \in \mathbb{R}^{[0,1]^N} \mid \exists \vec{p} \in \Delta_N \forall \vec{\omega} \in [0,1]^N \ g(\omega) = \vec{p} \cdot \vec{\omega}\}$ is obviously non-empty and it is compact and convex as a linear image of the simplex $\Delta_N$.

Now consider a function $H : \Gamma \times \Omega \to \mathbb{R}$ of the form

$$H(\gamma, \omega) = e^{\eta(\lambda(\gamma,\omega) - \lambda(\gamma',\omega)) - \eta^2/2}, \qquad (16)$$

where $\gamma' \in \Gamma$ and $\eta \geq 0$ are parameters.

**Lemma 8.** *Let $(\Omega, \Gamma, \lambda)$ be a game, the range of $\lambda$ be included in $[0, 1]$ and $G(\pi)$ be defined by (15). Then for all $\gamma' \in \Gamma$, for all $\eta \geq 0$, for all $\pi \in \mathcal{P}(\Omega)$, and for all $\gamma \in G(\pi)$ it holds that*

$$\mathrm{E}_\pi e^{\eta(\lambda(\gamma,\cdot) - \lambda(\gamma',\cdot)) - \eta^2/2} \leq 1.$$

The lemma follows from the Hoeffding inequality (see e.g. Cesa-Bianchi and Lugosi, 2006, Lemma A.1).

Now we can explain what $f_t$ is used in (14):

$$f_t(\gamma, \omega) = \sum_{k=1}^K p_{t,k} H_{t,k}(\gamma, \omega), \qquad (17)$$

where $p_{t,k} \geq 0$ are some weights and $H_{t,k}$ are functions of the form (16) with some parameters $\eta_{t,k}$ and $\gamma_{t,k}$, cf. (2), (5), and (7). The sum may be infinite or it can be even an integral over some measure (in place of weights $p_{t,k}$). As in the definition of supermartingale, the index $t$ may hide the dependence on a long sequence of arguments.

**Lemma 9.** *$S_t$ defined by (14), (15), and (17) satisfies the conditions of Lemma 5 if $(\Omega, \Gamma, \lambda)$ is a bounded convex compact game with finite $\Omega$ or the conditions of Lemma 6 if $(\Omega, \Gamma, \lambda)$ is DTOL, where $S_t(\pi)$ is taken for $q(\pi)$ and $\sum_{k=1}^K p_{t,k}$ is taken for $C$.*

### Acknowledgements

This work was supported by EPSRC, grant EP/F002998/1. We are grateful to Yura Kalnishkan for discussions.